# Resolving Class Imbalance in Object Detection with Weighted Cross Entropy Losses

**Trong Huy Phan**\*　　　　　　　　　　　　　　　　　　　　PHAN586@OKI.COM
**Kazuma Yamamoto**\*　　　　　　　　　　　　　　　　　YAMAMOTO377@OKI.COM
\**OKI Electric Industry Co., Ltd., 1-18-8 Chuou Warabi-shi, Saitama 335-8510, Japan.*

## Abstract

Object detection is an important task in computer vision which serves a lot of real-world applications such as autonomous driving, surveillance and robotics. Along with the rapid thrive of large-scale data, numerous state-of-the-art generalized object detectors (e.g. Faster R-CNN, YOLO, SSD) were developed in the past decade. Despite continual efforts in model modification and improvement in training strategies to boost detection accuracy, there are still limitations in detector's performance when it comes to specialized datasets with uneven object class distributions. This originates from the common usage of *Cross Entropy* loss function for object classification sub-task that simply ignores the frequency of appearance of object class during training, and thus results in lower accuracies for object classes with fewer number of samples. Class-imbalance in general machine learning has been widely studied, however, little attention has been paid on the subject of object detection. In this paper, we propose to explore and overcome such problem by application of several weighted variants of *Cross Entropy* loss, for examples *Balanced Cross Entropy*, *Focal Loss* and *Class-Balanced Loss Based on Effective Number of Samples* to our object detector. Experiments with BDD100K (a highly class-imbalanced driving database acquired from on-vehicle cameras capturing mostly "Car"-class objects and other minority object classes such as "Bus", "Person" and "Motor") have proven better class-wise performances of detector trained with the afore-mentioned loss functions.

## 1. Introduction

Object detection is one of the most fundamental and widely-studied tasks in computer vision community. It has been breaking into various industries with use cases ranging from image security, surveillance, automated vehicle systems to machine inspection. In this paper, we focus our discussion in object detection from vehicle mounted cameras which are becoming more and more common nowadays. Besides playing an extremely important role in autonomous driving, detected on-road objects such as cars, pedestrians from such cameras could also be further processed for number plate recognition, vehicle tracking, traffic condition monitoring and other useful applications.

In the past decade, with the help of large-scale database, numerous state-of-the-art deep learning-based generalized object detectors (e.g. Faster R-CNN (S. Ren et al., 2015), YOLO (J. Redmon et al., 2016), SSD (W. Liu et al., 2016)) were developed and currently widely used. Despite sequential successes in architecture design and training strategy which have led to remarkable improvements in overall detection accuracy for benchmarks such as PASCAL VOC (M. Everingham et al., 2015) and MS COCO (T. Lin et al., 2014), object detectors still face difficulties with specialized datasets having uneven object class distributions. The main reason behind such limitation is the common usage of *Cross Entropy* loss for object classification sub-task (X. Wu et al., 2020) that does not consider the frequency of appearance of object class during training, and therefore results in poor performance of object classes with fewer number of samples. Although there are several literature reviews on class-imbalance in general machine learning (e.g. M. Buda et al., 2018), to the best of our knowledge, very few have been made and examined on visual object detection task. Our main aim in this paper is to explore and overcome such problem by effective yet simple approach of applying weighted variants of *Cross Entropy* classification loss such as *Balanced Cross Entropy*, *Focal Loss* (T. Lin et al., 2017) and *Class-Balanced Loss Based on Effective Number of Samples* (Y. Cui et al., 2019) to the training of our object detector. The contribution of this paper is three-fold.

- First, we provide a brief review on class-imbalance in object detection task.
- Second, we propose to train our object detector with the afore-mentioned improved loss functions to tackle the class-imbalance problem.
- Third, we evaluate the class-wise detection performance of our proposal with Berkley DeepDrive 100K (BDD100K) (F. Yu et al., 2018), which is a highly class-imbalanced driving database acquired from on-vehicle camera capturing mostly "Car"-class object and other minority object classes such as "Bus", "Person" and "Motor".

## 2. Problem Formulation
### 2.1. Object Detection Task
Object detection task simply seeks the answers to i) where objects are located and ii) which category each object instance belongs to for every single image. Currently, deep learning-based object detection frameworks can be broadly divided into two groups: 1) Two-stage detectors, such as Region-based CNN (R-CNN) (R. Girshick et al., 2014) and its successors (i.e. R. Girshick, 2015; S. Ren et al., 2015) and 2) One-stage detectors, such as the YOLO family of detectors(J. Redmon et al., 2016-2018) and SSD (W. Liu et al., 2016).

As illustrated in Fig.1, two-stage detectors detect objects in two successive steps. Firstly, they use a region proposal generator to generate a set of object-like proposals and extract features from each proposal, which are then fed to a classifier that predicts the category of the proposed regions. On the other hand, one-stage detectors directly make categorical predictions of pre-defined proposals generated from each location of the feature maps regardless of whether it contains an object or a background region. Commonly, two-stage detectors achieve relatively better detection performance, whereas one-stage detectors are significantly faster and have greater applicability to real-time object detection.

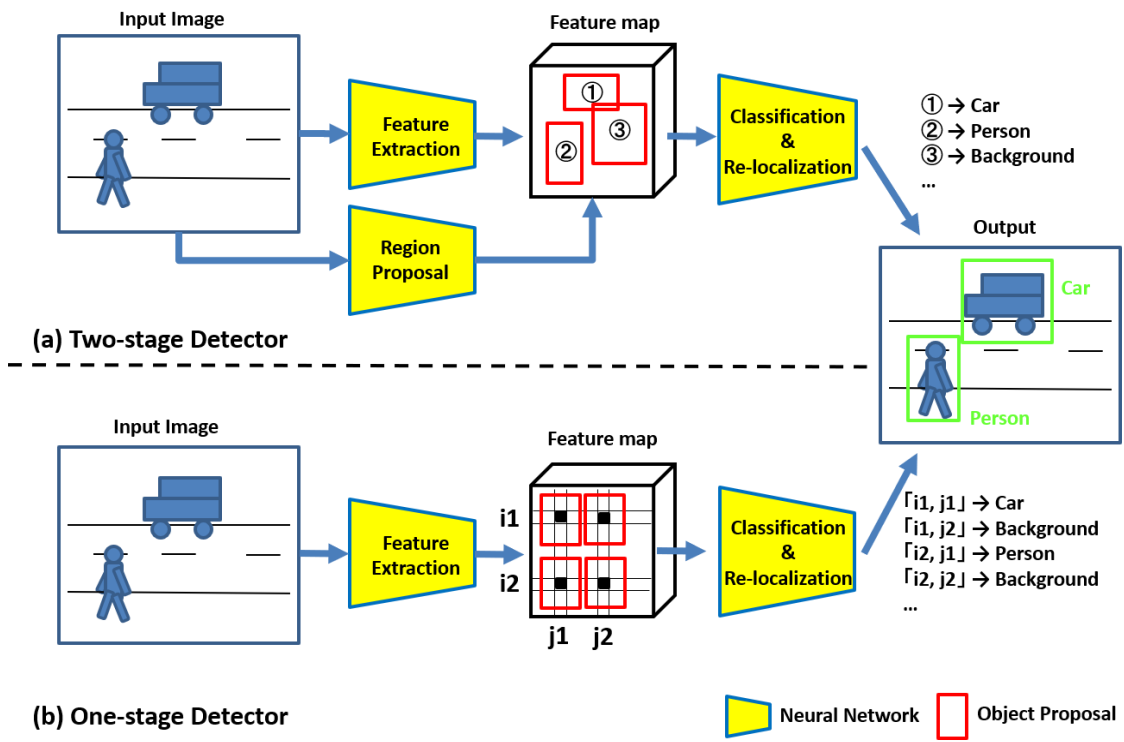

Fig. 1 Illustration of processing flows of (a)two-stage detector and (b)one-stage detector.

## 2.2. Class Imbalance in Object Detection

We argue that imbalance is observed during training of the above-mentioned proposal classifier of object detectors which would consequentially affect performance. Imbalance can occur in two fashions: foreground-objects-to-background imbalance and foreground-object-to-foreground-object imbalance.

Foreground-objects-to-background imbalance is inevitable in object detection task since most portion of the image is background. This type of imbalance has drawn a lot of research attention and is proven to be effectively alleviated by selectively sampling background regions during training (A. Shrivastava et al., 2016) to keep a fixed and balanced ratio of background to foreground objects.

Foreground-object-to-foreground-object imbalance is dataset dependent where certain object classes are over-represented compared to the others. In general, as the proposal classifier merely learns to minimize the overall classification error (i.e. *Cross Entropy* loss) accumulated from the training batch, under-represented object classes are prone to misclassification and lower detection accuracies. Although there are existing data-level methods to tackle the problem such as over-sampling of minority object classes or under-sampling of majority object classes (M. Buda et al., 2018), in this paper, we focus ourselves to cost-sensitive approach by simply applying different weighted versions of *Cross Entropy* classification loss which will be discussed in the following section.

## 3. Training Losses for Object Class Imbalance

In general, *Cross Entropy* is used to formularize the classification loss of deep learning-based object detectors (X. Wu et al., 2020). Given $C$-object classification task (where $C = \{0, 1, ..., C\}$; and $C=0$ indicates background class), *Cross Entropy* loss of the i-th object proposal can be calculated as:

$$\text{Loss}_{CE}(i) = - t_i \log(P(i)) \qquad (1)$$

In eq. (1), $t_i$ is the corresponding one-hot $C+1$-element vector indicating the ground-truth label. $P(i)$ is the confidence of the i-th object proposal belonging to each given class given by the object classifier which is commonly normalized with the softmax function or activated by a sigmoid function to transform into probability. This loss is afterward summed and averaged amongst all object proposals in the training batch to obtain the final classification loss for back propagation.

**Weighted Cross Entropy**. Since every object proposal is treated equally in the bare form of Cross Entropy loss (eq. (1)), object classes with few numbers of samples does not contribute significantly to the total loss and hence tend to be ignored during training. A very natural yet effective way to remedy this problem is to assign suitable weights to each object class which is represented in the following equation.

$$\text{Loss}_{WCE} = - w\, t_i \log(P(i)) \qquad (2)$$

$w$ in eq. (2) is a weight vector whose value can be user-chosen class by class and fixed or automatically adjusted during the training of object detectors. Note that $w$ simply equals to unity in the case of eq. (1). Larger value of $w$ increases the importance of the specified class during training. Next, we will introduce the weighting schemes of *Cross Entropy* loss which are to be evaluated in this paper.

**Balanced Cross Entropy**. As also mentioned by T. Lin et al. (2017), it is straightforward that a suitable fixed weight factor $w$ can be heuristically chosen by users. In this study, we simply assign $w = 1$ for background as well as majority object classes. For minority object classes, we set several arbitrary values of $w > 1$ and choose the one having the best performance in overall and class-wise accuracies.

**Inverse Class Frequency**. Another way to design suitable weight for each of object class is by inference from the inverse of number of samples which has been mentioned in the works of C. Huang et al. (2016), D. Mahajan et al. (2018) and Y. Wang et al. (2017). We observe that direct usage of inverse number or appearance frequency of object makes the training rather unstable as loss significantly rises when training batch contains extremely rare object classes. In this paper, we choose to adopt a linearly scaled representation $w = \frac{k}{f_j}$ and a logarithmic version $w = \log\left(\frac{q}{f_j}\right)$ of the inverse number of samples, where $f_j$ is the average number of samples of the j-th object class in a single image and $k > 0$; $q > f_j$ are hyper-parameters.

**Focal Loss**. Authors of the successful RetinaNet (T. Lin et al., 2017) have suggested automatic adjustment of the weight factor (also called "modulating factor" in the original paper) by using the confidence of object classifier $w = (1 - P(i))^\alpha$, where $\alpha$ is a real positive hyper-parameter. As the classifier is less confident in minority-class objects, lower value of $P(i)$ makes the correspondent weight $w$ larger and automatically bring focus to itself in the training of object classifier.

**Effective Number of Object Class**. Most recently, Y. Cui et al. (2019) has designed weighting scheme that uses the effective number of samples for each class to re-balance the classification loss. The effective number of samples is defined as the volume of samples and can be calculated by a simple formula given by $w = \frac{1-\beta^{n_j}}{1-\beta}$, where $n_j$ is the number of samples in the j-th object class and $\beta \in [0,1)$ is a hyper-parameter.

Although, experiments with weight combination (i.e. by multiplying the above-mentioned weight factors) is encouraged and could possibly bring about even better class-wise detection performances, it is beyond the scope of this paper.

## 4. Experiments

### 4.1. Dataset and Object Detector Description

To evaluate the effectiveness of weighted versions of *Cross Entropy* in imbalanced object detection task, we choose "BDD100K" which is currently known as the largest available dataset of annotated driving scenes, consisting of over 100K diverse video clips (F. Yu et al., 2018). Compared to other common datasets, BDD100K captures more of the "long-tail" of object class variation (Fig. 2) in diverse environmental domains that we find suitable for our study. Among the publicly available annotated image sets, we split the Training-set into 60,000 and 10,000 images for training and validation respectively; and use the 10.000 images in Validation-set for evaluation. Seven on-road objects classes are targeted in our experiments, namely "Car", "Truck", "Bus", "Person", "Rider", "Motor" and "Bike", among which "Car" is the most frequently presented object class compared to the others.

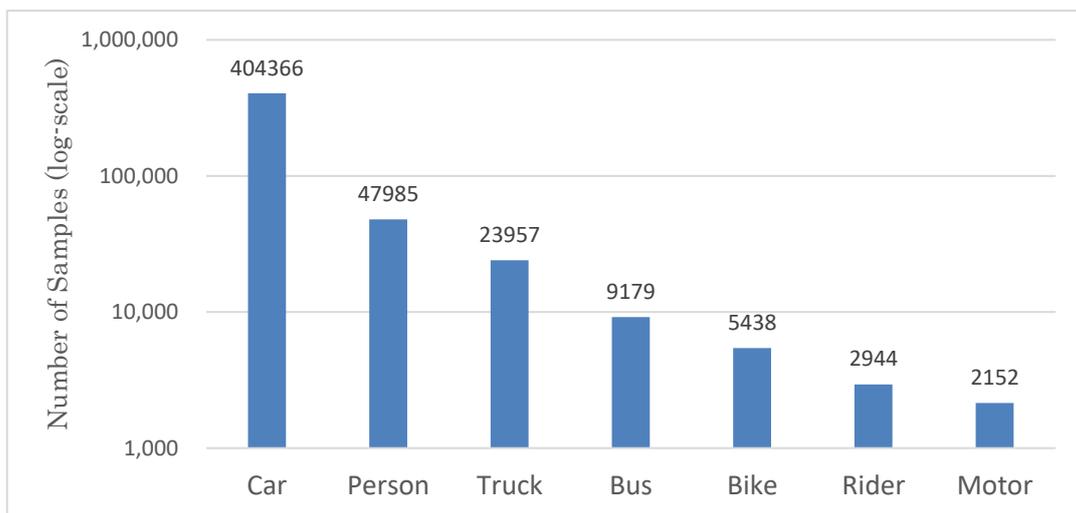

Fig. 2 Number of samples (log-scale) of each object class in BDD100K (Training-set and Validation-set only).

As for the choice of object detector, we adopt the design of one-stage detector SSD (W. Liu et al. 2016) and add several up-sampling convolution layers to the architecture (e.g. L. Cui et al., 2018; C. Fu et al., 2017) to make it more robust to the frequently appeared small objects. VGG16 (K. Simonyan et al. 2014) is used as the backbone of our model. Since there is no specific assumption should be made on choice of detector architecture, we are confident that the findings in this paper are also applicable to other object detectors as well as other databases.

In all experiments, we train our detector with Momentum SGD. Classification training are performed with hard negative mining (A. Shrivastava et al., 2016) such that the foreground objects to background ratio is 1:3 which leads to a faster and more stable training. Probability of an object belonging to each class is softmax-normalized. Furthermore, to make the model more robust to variance in appearance, we augment training images with flipping, cropping, color distortions etc. as originally mentioned in W. Liu et al., (2016).

### 4.2. Evaluation Metric

Mean Average Precision (mAP) is usually the choice for performance comparison of object detectors on major benchmark such as MS COCO (T. Lin et al., 2014). In this paper, to evaluate balance in class-wise performance, we set the same score threshold for all object class and use recall under fixed False Positive Per Image (FPPI). In addition, a proposed object is deemed to be True Positive if and only if its IoU (Intersection over Union) with the ground-truth box is larger than a predefined threshold and the inferred object class matches the ground-truth label. Throughout this paper FPPI = 1 and IoU = 0.5 are used.

### 4.3. Effectiveness of Different Weighted Cross Entropy Losses for Object Class Imbalance

Table 1. summarizes the evaluation results of different weighted *Cross Entropy* losses. The chosen hyper-parameters and their corresponding weights are tabulated in Table 2.

Table 1. Class-wise recall (FPPI = 1) of different weighted *Cross Entropy* losses.

| Object Class | Original Cross Entropy | Focal Loss | Balanced Cross Entropy | Inverse Class Frequency | | Effective Number of Samples |
| | | | | Linear | Logarithmic | |
|---|---|---|---|---|---|---|
| Bike | 19.10% | 30.80% | 44.80% | **57.70%** | 56.40% | 49.10% |
| Bus | 45.10% | 53.20% | 59.70% | 66.50% | **66.70%** | 60.70% |
| Car | 81.00% | **81.60%** | 77.00% | 78.50% | 78.30% | 75.70% |
| Motor | 17.20% | 31.00% | 42.90% | **67.10%** | 58.00% | 57.70% |
| Person | 35.60% | 42.00% | **65.00%** | 38.20% | 39.70% | 62.20% |
| Rider | 44.50% | 51.10% | **68.20%** | 38.60% | 41.60% | 65.80% |
| Truck | 49.20% | 52.80% | **64.00%** | 57.70% | 63.60% | 62.50% |
| Average | 41.70% | 48.90% | 60.20% | 57.80% | 57.70% | **61.90%** |
| Overall | 73.20% | **74.90%** | 74.30% | 72.90% | 73.00% | 73.00% |

Table 2. Chosen hyper-parameters and their corresponding weights.

|  |  | Back-ground | Car | Truck | Bus | Person | Motor | Bike |
|---|---|---|---|---|---|---|---|---|
| Original Cross Entropy | | 1 | 1 | 1 | 1 | 1 | 1 | 1 |
| Balanced Cross Entropy | | 1 | 1 | 5.00 | 5.00 | 5.00 | 5.00 | 5.00 |
| Inverse Class Frequency | Linear (k=0.5) | 1 | 1 | 2.92 | 7.63 | 1.37 | 32.53 | 12.87 |
| | Logarithmic (q=20) | 1 | 1 | 4.66 | 8.82 | 1.38 | 15.12 | 11.10 |
| Focal Loss (α=2) | | - | - | - | - | - | - | - |
| Effective Number of Samples (β=0.9) | | 1 | 1 | 5.14 | 6.68 | 5.00 | 17.95 | 8.93 |

All weighted versions of *Cross Entropy* showed better performance in minority object classes such as "Person", "Truck", "Bus" which gave rise to a maximum of roughly 20% in class-average recall. Compared to the conventional unweighted *Cross Entropy*, even though there was a slight drop in detection performance of the majority object class "Car" since classification training favored it less, the overall recall was not significantly affected (Fig. 3).

In terms of balance in class-wise detection performance, weighting by *Effective Number of Samples* had the best result followed by *Balanced Cross Entropy* and by using *Inverse Class Frequency*. As anticipated, weighting by *Inverse Class Frequency* boosted the recall of super rare object classes such as "Bike" and "Motor" to a great amount, but adversely worsened the performance of classes having relatively higher number of samples such as "Person". On the other hand, our chosen weight factor for *Balanced Cross Entropy* showed comparable result to that of *Class-Balanced Loss Based on Effective Number of Samples* without hurting the overall performance.

*Focal Loss* performed least satisfactorily with the rise of less than 10% in average recall yet achieved the best overall recall. This can be argued by the fact that class weight (modulation factor) in *Focal Loss* does not depend on prior knowledge of majority or minority in object classes. Besides "car", background class accounts for most of the samples in training batch (i.e. three times the number of foreground objects in our experiments). Thus, Focal Loss tends to optimize the imbalance in classification in the foreground-objects-to-background direction which explains the highest overall recall.

In summary, class-weighting of *Cross Entropy* loss in any form can effectively improve the detection performance of minority object classes. Although our experiments have shown that the level of balance between object classes is governed by the choice of weight factor to some extends, it is possible to fine-tune the class-wise performance according to specific applications by combining multiple training stages for object classifier.

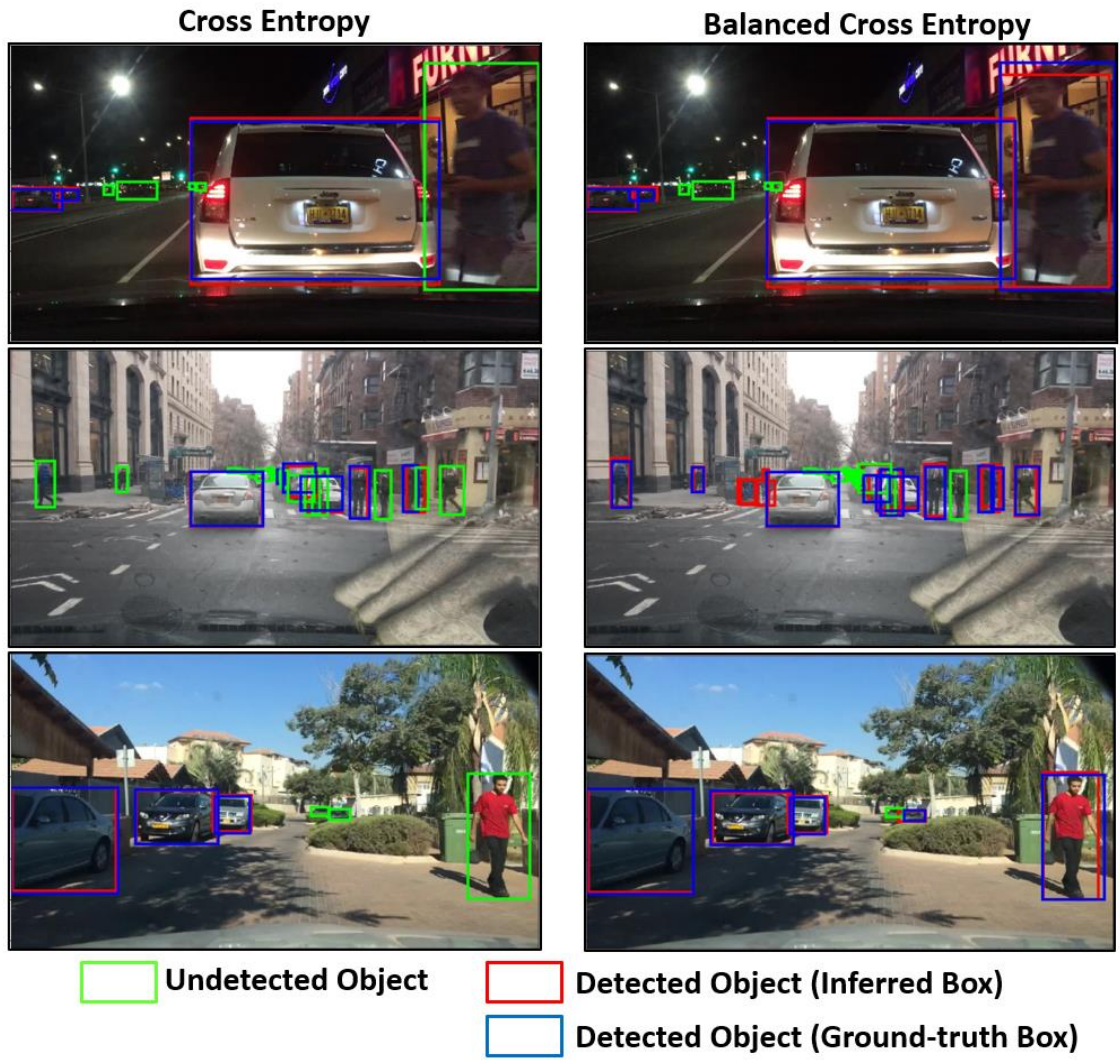

Fig. 3 Visualization of detection results for detectors trained with the *Cross Entropy* loss (left column) and Balanced Cross Entropy loss (right column). Without suitable weighting, the conventional *Cross Entropy* misses a lot of "Person" instances.

To further improve detection accuracies of minority object classes, our analysis shows that data related problems need to be addressed as well. For examples, in BDD100K dataset, bounding box annotation of "Motor" or "Bike" objects sometimes captures mainly the rider-region which is undesirable when training together with ordinary "Person" class. In addition, vehicle object classes namely "Car", "Truck" and "Bus" contain a lot of false annotations and mutually share van-like vehicle (Fig. 4) which affect training and consequentially lead to mis-classification among such object classes. The problems mentioned here could be lifted by data-cleansing or by designation of soft target label (R. Müller et al., 2019) for specified object classes which we hope to consider in our future works.

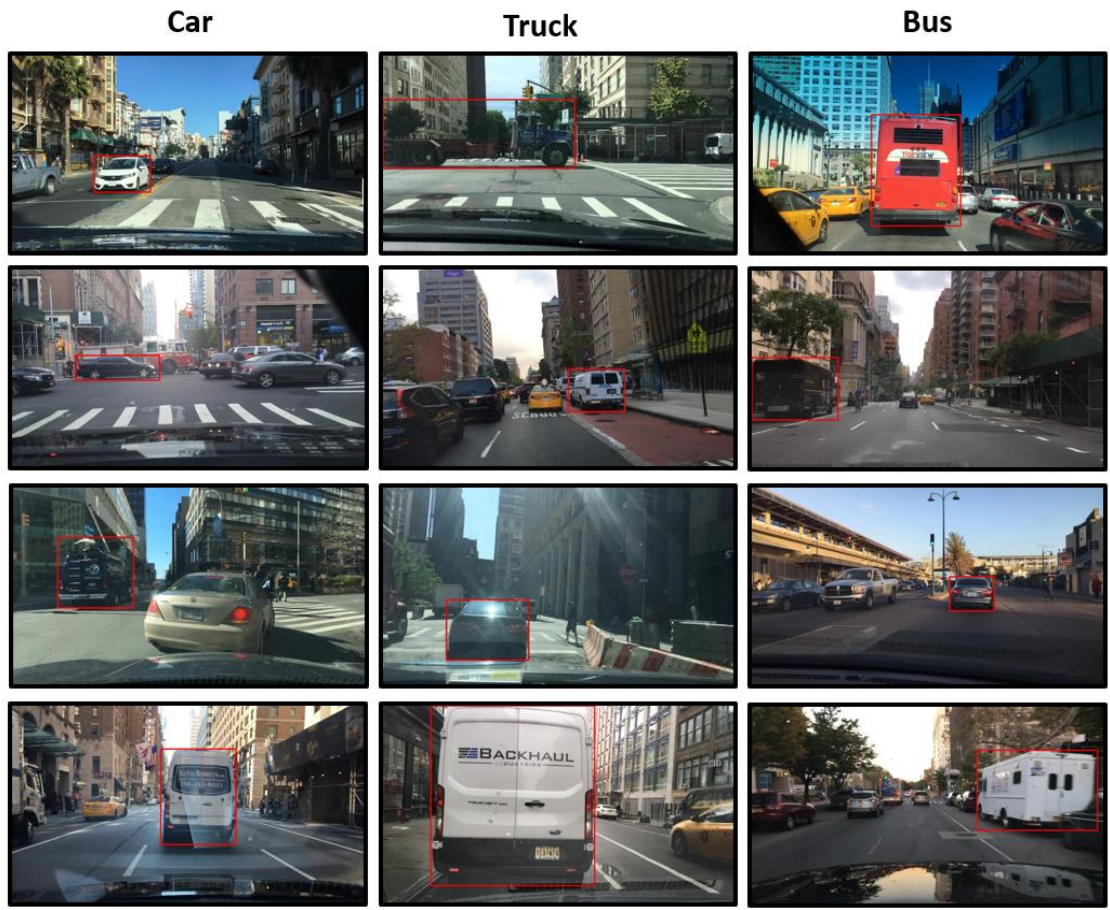

Fig. 4 Examples of annotations for "Car" (1st column), "Truck" (2nd column) and "Bus" (3rd column). They mutually contain van-like object instances. Besides, a number of "Truck" and "Bus" samples are actually cars.

## 5. Conclusion

In this paper, we have provided a brief review on class-imbalance in object detection task and our proposal to counter the problem with the application of different variants of weighted *Cross-Entropy* loss. Our experiments with BDD100K driving database have shown noticeable improvements in class-wise detection accuracy as compared to the commonly used unweighted counterpart. Although we have limited ourselves to specific dataset and choice of object detector in this study, we are confident that the findings in this paper are broadly applicable. Furthermore, re-weighting of classification Cross Entropy loss could certainly be extended to other aspects of imbalances in object detection such as object sizes, variants in appearances, visual domains etc. which we think is worthy to explore in the future.

**Acknowledgments**

Shaoqing Ren, Kaiming He, Ross Girshick, and Jian Sun. Faster r-cnn: Towards real-time object detection with region proposal networks. In: *Advances in Neural Information Processing Systems*. 2015. p. 91-99.

Yu-Xiong Wang, Deva Ramanan, and Martial Hebert. Learning to model the tail. In: *Advances in Neural Information Processing Systems*. 2017. p. 7029-7039.

Xiongwei Wu, Doyen Sahoo, and Steven C.H. Hoi. Recent advances in deep learning for object detection. In *Neurocomputing*, 2020.

Fisher Yu, Haofeng Chen, Xin Wang, Wenqi Xian, Yingying Chen, Fangchen Liu, Vashisht Madhavan, and Trevor Darrell. Bdd100k: A diverse driving video database with scalable annotation tooling. *arXiv preprint arXiv*:1805.04687, 2018.